\documentclass[10pt,twocolumn,letterpaper]{article}

\usepackage{cvpr}
\usepackage{times}
\usepackage{epsfig}
\usepackage{graphicx}
\usepackage{amsmath}
\usepackage{amssymb}
\usepackage{comment}
\makeatletter
\newcommand{\printfnsymbol}[1]{%
  \textsuperscript{\@fnsymbol{#1}}%
}
\makeatother

\usepackage[pagebackref=true,breaklinks=true,letterpaper=true,colorlinks,bookmarks=false]{hyperref}

\cvprfinalcopy 

\def\cvprPaperID{****} 

\ifcvprfinal\fi
\begin{document}

\title{Cycled Compositional Learning between Images and Text}

\author{Jongseok Kim$^1$\thanks{equal contribution} \hspace{9pt} \ Youngjae Yu$^{1,2}$\printfnsymbol{1} \hspace{9pt}  Seunghwan Lee$^2$ \hspace{9pt}  GunheeKim$^{1,2}$\\
Seoul National University$^1$, RippleAI$^2$ \\
Seoul, Korea\\
{\tt\small js.kim@vision.snu.ac.kr, \{yj.yu,seunghwan,gunhee\}@rippleai.co}
}

\maketitle

\begin{abstract}

We present an approach named the \textit{Cycled Composition Network} that can measure the semantic distance of the composition of image-text embedding. 
First, the \textit{Composition Network} transit a reference image($RefImg$) to target image($TrgImg$) in an embedding space using relative caption($Text$).
Second, the \textit{Correction Network} calculates a difference between reference and retrieved target images in the embedding space and match it with a relative caption.
Our goal is to learn a \texttt{Composition} mapping with the \textit{Composition Network}  $G: (RefImg, Text) \rightarrow TrgImg$. Since this one-way mapping is highly under-constrained, we couple it with an inverse relation learning with the \textit{Correction Network} $F: (RefImg, TrgImg) \rightarrow Text$ and introduce a cycled relation  $F(RefImg,G(RefImg,T)) \approx T$ for given $RefImg,T$.

We participate in Fashion IQ 2020 challenge~\cite{fashioniq-arxiv-2019} and have won the first place with the ensemble of our model. 
The code is available at \url{https://github.com/ozmig77/CCnet_fashioniq2020}.
\end{abstract}

\section{Introduction}
%

\begin{figure}[t]
\centering
\includegraphics[trim=0.0cm 0.0cm 0cm 0.0cm,clip,width=0.47\textwidth]{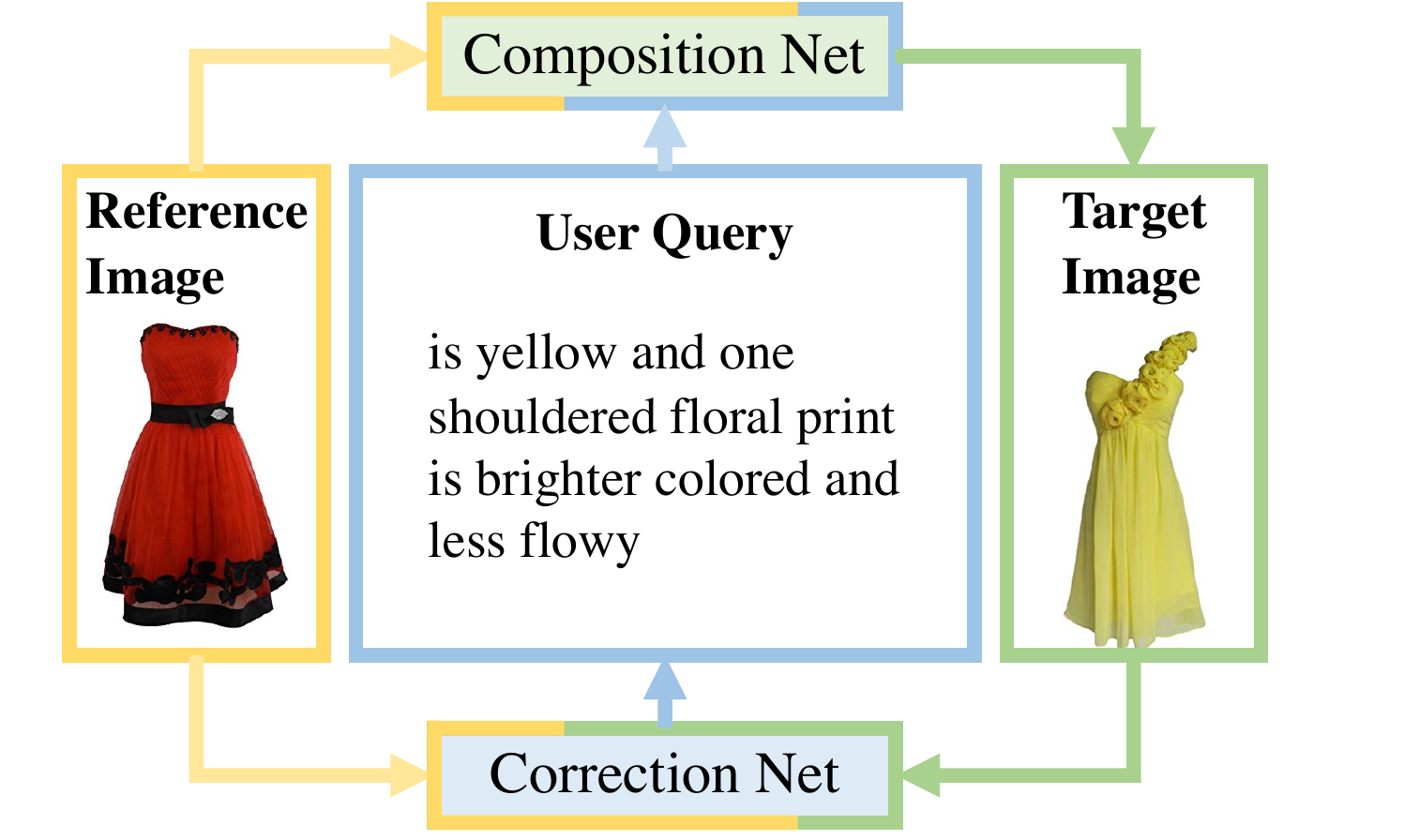}
\caption{
The key intuition of the \texttt{Cycled Composition} model. 
Given a triplet of (reference image, Language query, Target image), 
Our model effectively encodes the differential relationship between the two images in both Forward (Composition Net) and Inverse (Correction Net) pathways.
}
\label{fig:keyidea}
\end{figure}
For implementing interactive conversational image search systems, it is a prerequisite to develop a controllable image search based on user feedback (\eg natural language queries and/or category filters). Beyond its practical and commercial consequence that can promote users' convenience of shopping significantly, it is also an intriguing research problem to study the multimodal embedding and cross-modal retrieval.

In this work, we propose an approach that can measure the semantic similarity between the composition embedding of images and text,
and apply it to tackle controllable multimodal image retrieval for fashion data.
We summarize the contributions of this work as follows:
\begin{enumerate}
\item We propose the \textit{Cycled Composition Network}, which can measure the semantic differential relationship between images for a query relative caption. Compare to existing models such as TIRG~\cite{vo-cvpr-2019}, CurlingNet~\cite{yu2020curlingnet} our model not only learns the image-text composition features for finding the most suitable target data but also focuses on the queried attributes in target data for better ranking.
\item To validate our proposed model, we participate in the Fashion-IQ 2020 challenge and our ensemble model achieves the first place on the leaderboard. 
\end{enumerate}


\section{Approach}

In the Fashion-IQ challenge, a participant needs to retrieve the best matching target image, given reference image and relative caption.
Since attributes of fashion correspond to various parts of the image, we utilize a mixture of experts to solve the task.

Besides, we propose a Correction Network that find the difference between reference and target images and checks its validity with a relative caption.

\subsection{Image Experts}
For getting fashion related features from image, we first pre-train deep CNNs(\eg ResNet-152~\cite{he-arxiv-2015}, DenseNet-169 ~\cite{huang-cvpr-2017}) with Deepfashion\cite{liu-cvpr-2016} attribute prediction task. And use the CNNs  as a basic backbone of our model.

To spot out fashion attributes in diverse aspect, in addition to use global pooled embedding from last layer of pre-trained CNN, we also utilize output of intermediate layer and 3 x 3 features slice from multi locations([0:3,2:5], [2:5,0:3], [2:5,2:5], [2:5,4:7] and [4:7,2:5] from the $7\times7$) follow by average pooling.

Then we pass embeddings into expert specific fully connected layer follow by context gating\cite{miech-arxiv-2017} and get 7 different image expert embeddings.

We share parameters for image experts between reference and target images.
In the end, we get reference and target image embedding for each expert $e$, $x^{ref}_e, x^{trg}_e \in \mathbb{R}^{D}$ where $D$ is a hidden dimension.

\subsection{Text Experts}

Given relative caption with $l$ words, We first embed word using GloVe\cite{Pennington-emnlp-2014}, $w^* = [w^*_1, ..., w^*_l] \in \mathbb{R}^{l \times 300}$.
Then we get caption embedding $w \in \mathbb{R}^{l \times D}$ by,
\begin{align}
\label{eq:wordembedding}
    w = \text{FC}([\text{Conv1d}(w^*) ; w^*])
\end{align}
where Conv1d is a convolution 1d layer, FC is a fully connected layer, and $[;]$ is a concatenation. 

Each image expert has different words to be focused on.
For example, in Figure~\ref{fig:keyidea} it is natural that experts with global information attend to "yellow" or "less flowy" while the expert with  the upper part attends to "shouldered floral print". 
To take advantage of it, we assign randomly initialized characterize embedding $m_e \in \mathbb{R}^{D}$ for each expert $e$. Then perform attended pooling with caption embedding  $w$ to get attended embedding $t^*_e \in \mathbb{R}^{D}$,
\begin{align}
    \alpha_e &= soft\max_l(\text{FC}(\text{FC}(m_{e} \odot w))) \\
    t^*_e &= \sum_l \alpha_{e,l} w_{l}
\end{align}
where $\odot$ is hadamard product.

Finally we further embed $t^*_e$ with expert specific fully connected layer follow by context gating\cite{miech-arxiv-2017} and get text expert embedding $t_e$.

\subsection{Cycled Composition Networks}
\textbf{Composition Network}
Composition Network finds the best matching target image from a given reference image and relative caption.
To compose image and text, we use gating module intuited by Text Image Residual Gating (TIRG)~\cite{vo-cvpr-2019} to manipulate the vector transition.

\begin{align}
    \bar{t}_e &= [t_e; \text{Fusion}(x^{ref}_e, t_e)] \\ 
    c_e &= w_gf_{gate}(x^{ref}_e, \bar{t}_e) + w_rf_{res}(x^{ref}_e, \bar{t}_e)
\end{align}
where $w_g, w_r$ are learnable paramters, and we use MUTAN~\cite{mutan-iccv-2017} for fusion function. The gating and residual functions are computed by,
\begin{align}
    f_{gate}(x^{ref}_e, \bar{t}_e) &= \sigma(\text{FC}(\text{FC}([x^{ref}_e; \bar{t}_e])) \odot x^{ref}_e \\
    f_{res}(x^{ref}_e, \bar{t}_e) &= \text{FC}(\text{FC}([x^{ref}_e; \bar{t}_e]))
\end{align}
where $\sigma$ is the sigmoid function.

Then we calculate matching score $s^{r}$ by, 
\begin{align}
    s^{r}(\textbf{x}^{ref},\textbf{t},\textbf{x}^{trg}) = \sum_e c_e \cdot x^{trg}_e 
\end{align}
where $\cdot$ is dot product.
In minibatch size of $B$ consist of ground truth triplets $\{\textbf{x}^{ref}_i,\textbf{t}_i,\textbf{x}^{trg}_i\}_{i=1}^B$, we use loss function object to find best matching target image given reference image and text. 
\begin{align}
\label{eq:loss}
    L = \frac{1}{B} \sum^B_{i=1}-\log \frac{exp(s^{r}(\textbf{x}^{ref}_i,\textbf{t}_i,\textbf{x}^{trg}_i))}
                         {\sum^B_{j=1}exp(s^{r}(\textbf{x}^{ref}_i,\textbf{t}_i,\textbf{x}^{trg}_j))}
\end{align}

\textbf{Correction Network}
Correction Network verifies that the difference between reference and target image explains the relative caption. 
By this, we double check that the target image is related to the reference image and the text.

To represent the difference between reference and target image, we use the following approach,
\begin{align}
    \bar{x}^{trg}_e &= FC([x^{trg}_e \odot x^{ref}_e ; x^{trg}_e])\\
    \bar{x}^{ref}_e &= FC([x^{trg}_e \odot x^{ref}_e ; x^{ref}_e])\\
    \bar{x}^{diff}_e &= \bar{x}^{trg}_e - \bar{x}^{ref}_e \\
    d_e &= \text{FC}([x^{ref}_e;x^{trg}_e;\bar{x}^{diff}_e])
\end{align}
The intuition is that concatenate with fused embedding before subtraction allows features to know what to subtract.

Then we calculate the matching score by,
\begin{align}
    s^{c}(\textbf{x}^{ref},\textbf{x}^{trg},\textbf{t}) = \sum_e d_e \cdot t_e
\end{align}


Training is done by predicting the best matching target image with a ground truth reference image and relational text which identical to Eq.\ref{eq:loss}.

\textbf{CCNet}
Utilize results from both Composition Network and Correction Network we calculate the final probability by,
\begin{align}
    p(\textbf{x}^{trg}|\textbf{x}^{ref},\textbf{t}) = 
    p^r(\textbf{x}^{trg}|\textbf{x}^{ref},\textbf{t})
    p^c(\textbf{x}^{trg}|\textbf{x}^{ref},\textbf{t})
\end{align}
where $p^r$ and $p^c$ are probability obtain by softmax of score $s^r$ and $s^c$ from Composition Network and Correction Network. 

\begin{figure}[t]
\centering
\includegraphics[trim=0.0cm 0.0cm 0cm 0.0cm,clip,width=0.47\textwidth]{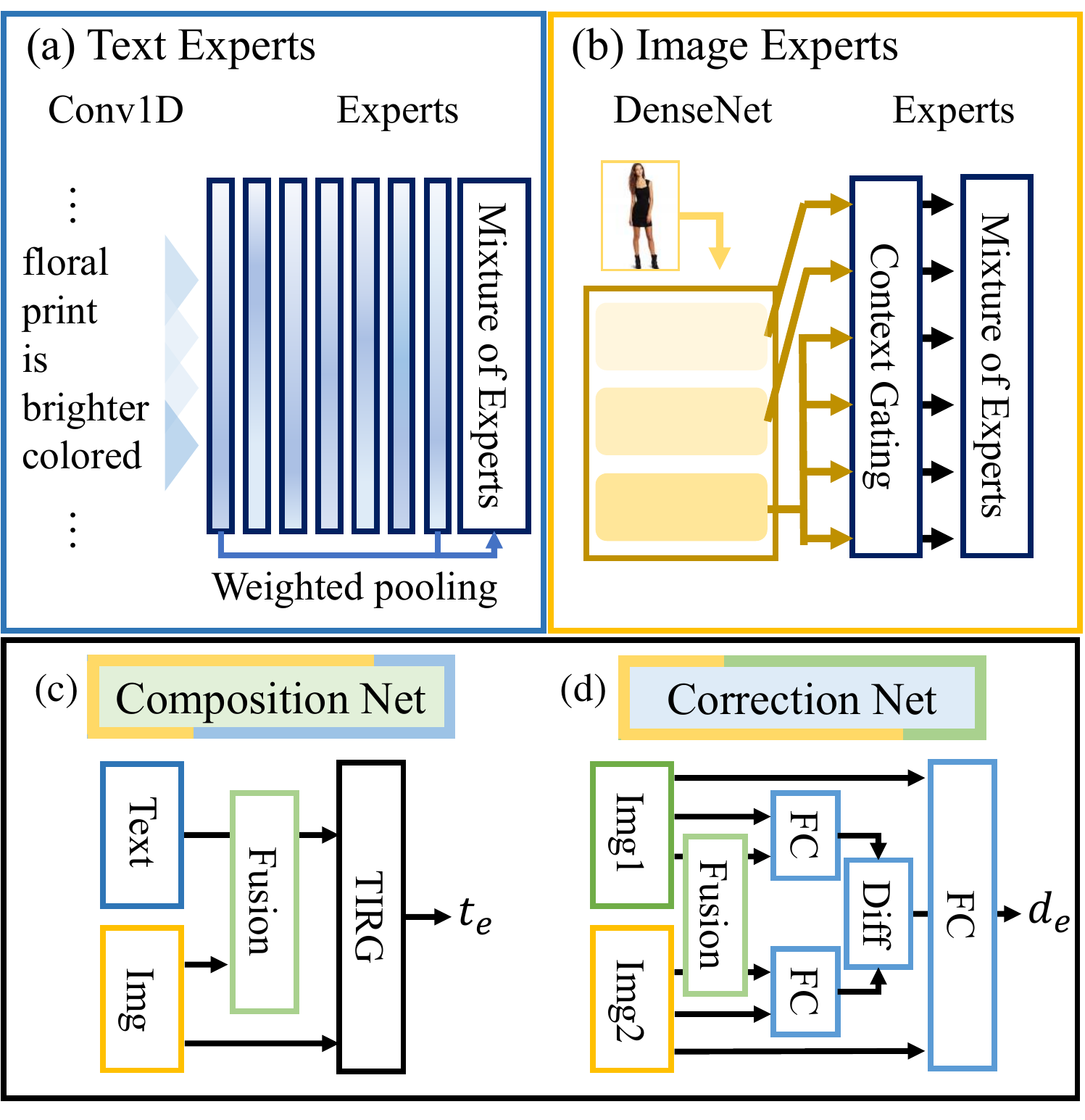}
\caption{
Schematics for the Composition and the Correction Networks. Best viewed in color.
}
\label{fig:joint_detail}
\end{figure}
\begin{table*}[t]
\begin{center}
 \begin{tabular}{|l | c c | c c | c c | c|} 
 \hline
 \multicolumn{1}{|c|}{Category} & \multicolumn{2}{|c|}{Dress} & \multicolumn{2}{|c|}{Shirt} & \multicolumn{2}{|c|}{Toptee} & Total \\
 \hline
 \multicolumn{1}{|c|}{Metric} & R@10 & R@50 & R@10 & R@50 & R@10 & R@50 & Avg \\ [0.5ex] 
 \hline\hline
 single conv1d          & 0.2716 & 0.5488 & 0.2227 & 0.4563 & 0.3059 & 0.5690 & 0.3957 \\ 
 multi conv1d           & 0.3024 & 0.5434 & 0.2277 & 0.4652 & 0.3274 & 0.5834 & 0.4082 \\
 shared fc              & 0.2737 & 0.5335 & 0.2066 & 0.4116 & 0.3004 & 0.5528 & 0.3847 \\
 \hline
 CurlingNet~\cite{yu2020curlingnet} & 0.2615 & 0.5324 & 0.2145 & 0.4456 & 0.3012 & 0.5523 & 0.3845 \\ 
 CompositionNet         & 0.3044 & 0.5741 & 0.2375 & 0.4701 & 0.3146 & 0.5762 & 0.4128 \\
 CorrectionNet          & 0.3153 & 0.5811 & 0.2537 & 0.4912 & 0.3345 & 0.5926 & 0.4281 \\ CCNet                 & \textbf{0.3456} & \textbf{0.6163} & \textbf{0.2767} & \textbf{0.5147} & \textbf{0.3677} & \textbf{0.6313} & \textbf{0.4587} \\
 \hline
 Ensemble (val split)   & 0.4075 & 0.6748 & 0.3229 & 0.5687 & 0.4202 & 0.6818 & 0.5126 \\ 
 Ensemble (test split)  & 0.4289 & 0.6749 & 0.3376 & 0.5790 & 0.4222 & 0.6702 & 0.5188 \\  
 \hline
\end{tabular}
\vspace{8pt}
\caption{\label{tbl:results}The qualitative result on Fashion-IQ dataset validation split. Top: ablation study of text experts model. Middle: comparison with baselines and the Cycled Composition Networks. Bottom: the result of ensemble models. $^\ddagger$: based on an official leaderboard results. 
Rounded up to 2 decimals }
\end{center}
\end{table*}

\section{Experiment}

In this section, we report the experimental results of CCNet for the fashion-IQ challenge.
The challenge provides an image retrieval dataset where the input query is specified in the form of a candidate image and two natural language expressions that describe the visual differences of the search target.

The dataset contains 77,683 fashion images and 30,134 pairs of relative captions for interactive image retrieval.
The participant needs to retrievthe best matching image given a pair of a reference image and relative caption.

\textbf{Evaluation Metrics.}
Every participant is evaluated by the recall metrics on the test splits of the dataset. 
For each of the three fashion categories (dresses, tops, shirts), Recall@10, and Recall@50 are computed on all test queries.
The overall performance is evaluated based on the average of Recall@10 and Recall@50. 
We strictly follow the evaluation protocols of the challenge\footnote{\url{https://sites.google.com/view/cvcreative2020/fashion-iq}}.


\subsection{Training Details}

In our main model, we use DenseNet-169\cite{huang-cvpr-2017} for backbone of image embedding.
For image augmentation, we use random crop and random horizontal flip.
To get the GloVe vector, we use spacy en\_vectors\_web\_lg.
The ReLU activation layer is used in every FC layer, and the hidden dimension $D$ is set as 1024.
For regularization, we apply batch normalization~\cite{Sergey-icml-2015}, and use dropout~\cite{srivastava-jmlr-2014} with probability 0.2.

Each training batch consists of $B=32$ triplets of (reference image, query text, target image).
We use batch shuffling in every training epoch.
We use Adam\cite{kingma-iclr-2015} optimizer with learning rate 1e-4 with the exponential learning rate decay of 0.95 per step.

\textbf{Ensemble.}
For final ensemble, we use several variants of
1. backbone(ResNet50, ResNet152, FishNet99\cite{sun-neurips-2018})
2. text embedding(Word2Vec, BERT)
3. text encoder(single conv, multi conv, FC, average pooling)
4. video experts (spatial experts, multi global pooled experts)
5. composition method (TIRG, FiLM, MUTAN, concat)
6. data augmentation (affine transformation, color jittering, back translation)
7. hidden dimension size, batch size, random seeds

\subsection{Quantitative Results}
Table~\ref{tbl:results} summarize the quantitative results of our experiments.
The top part shows ablation study of text experts model in Composition Network.
Where single conv1d uses the same average pooled text embedding for all image experts,
multi conv1d assigns independent text experts to each image experts i.e. parameters are not shared in word embedding step(Eq.\ref{eq:wordembedding}),
shared fc uses FC layer instead of Conv1d
and CompositionNet is our proposed text expert model.
As in the table, we show that our designed text expert model achieves better performance than other methods.

The middle part shows comparison with baselines and the Cycled Composition Networks.
As in the table, our Composition Network outperforms the results of last year's participants.
Furthermore, we show that Correction Network solely outperforms the conventional Retrieval setting.
Finally utilizing both networks(CCNet) shows a better result.

As a final result, our ensemble model achieves a 0.5126 average recall in validation split and 0.5188 in the test split.

\subsection{Qualitative Results}
\begin{figure}[t]
\centering
\includegraphics[trim=0.0cm 0.0cm 0cm 0.0cm,clip,width=0.47\textwidth]{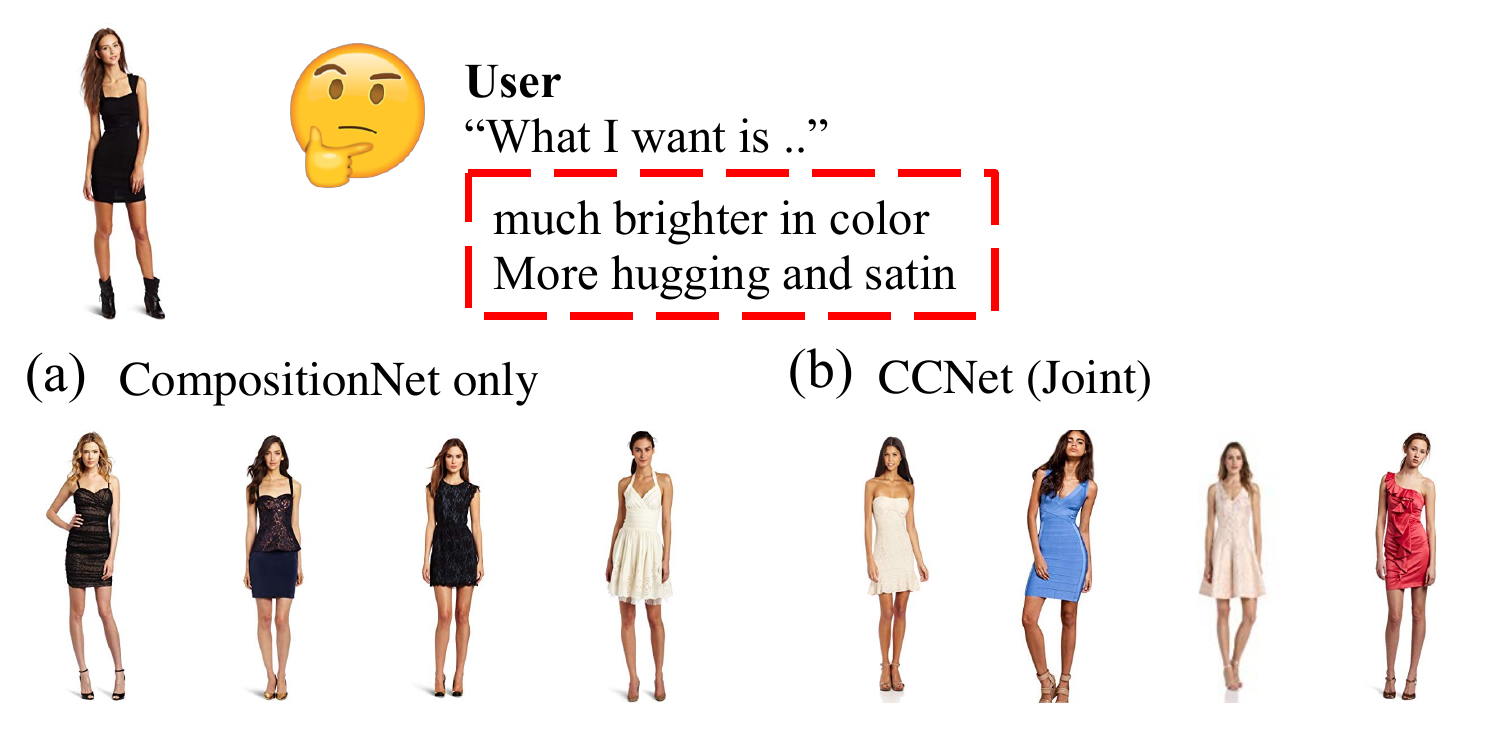}
\caption{Qualitative examples of the CCNet compared to ablational variant, CompositionNet only.
}
\label{fig:examples_ablation}
\end{figure}

\begin{figure}[t]
\centering
\includegraphics[trim=0.0cm 0.0cm 0cm 0.0cm,clip,width=0.47\textwidth]{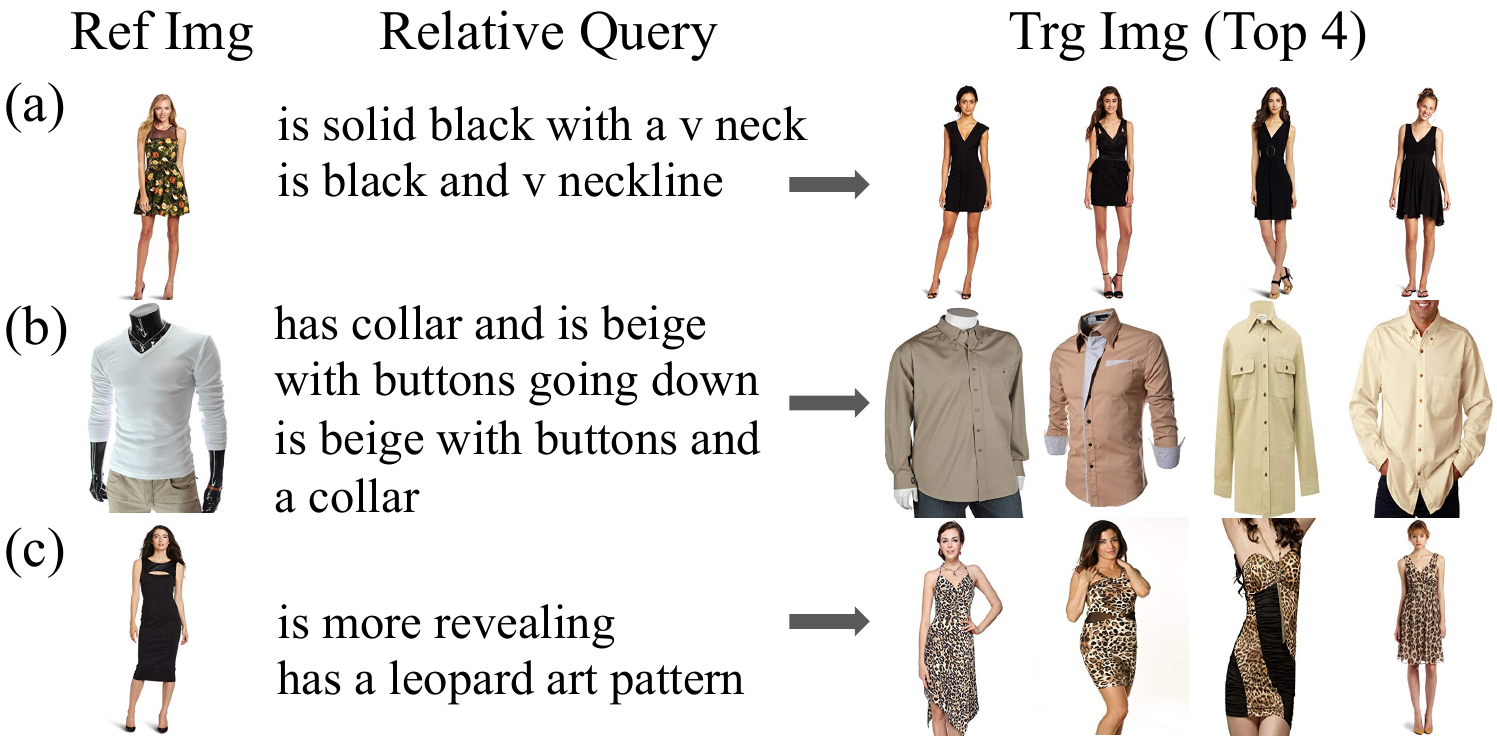}
\caption{Qualitative examples of the image-text retrieval task in the Fashion-IQ dataset.
}
\label{fig:examples}
\end{figure}

Figure \ref{fig:examples_ablation} shows difference between our joint model and classic CompositionNet. Thanks to double checking by correction network, our joint method find clothes that satisfy all conditions in the caption. 
Figure \ref{fig:examples} illustrates the qualitative results of our method.
We display reference images and text queries (Left) and the highest-scored retrieved images (Right) for some test examples.
While maintaining the style of the reference image, our model assigns a high score to the sample that well reflects the needs of the user query.

\begin{table}
\begin{center}
\begin{tabular}{|l|l|c|}
\hline
Submission       & Team Name  &{\footnotesize Avg R@(10,50)}   \\ \hline\hline
cuberick          & CARDINAL   & 0.33   \\
emd               &            & 0.39   \\
skywalker~\cite{li2019designovels}   & DESIGNOVEL & 0.44   \\
shuan~\cite{yu2020curlingnet}        & RippleAI   & 0.47   \\
superraptors      &            & 0.49   \\ \hline
zyday             &            & 0.43   \\
stellastra        & VAA        & 0.48  \\
ganfu.wb          &            & 0.50  \\
tianxi.tl         & Alibaba    & 0.50   \\ \hline
Ours              & RippleAI   & \textbf{0.52}  \\
\hline
\end{tabular}
\end{center}
\medskip
\caption{
Performance comparison for final submission on Test Phase. 
We also report the results of last year's challenge (FashionIQ 2019) in the top part of the table.
This year, all participants are not allowed to use attribute metadata for test split, which is crucial to improve performance in last year's challenge.
We finished the first place in the challenge.
}
\vspace{-3pt}

\label{tbl:results_ensemble}
\end{table}

\section{Conclusion}

We proposed the Cycled Composition Network(CCNet) for learning differential relations between two image embeddings with respect to a text query.
The two key components of the model Composition, and Correction networks are easily adaptable in many image-text composition tasks, including user adjustable retrieval or image recommendation systems.
We demonstrated that our method improved the performance of image-text composition learning.

Moving forward, we plan to expand the applicability of the Cycled Composition network; we can tackle relative caption/dialogue generation task for multimodal conversational systems.

{\small
\bibliographystyle{ieee_fullname}
\bibliography{paper-cvpr20-fashion}

\begin{thebibliography}{10}\itemsep=-1pt

\bibitem{mutan-iccv-2017}
Hedi Ben-Younes, R{\'e}mi Cadene, Matthieu Cord, and Nicolas Thome.
\newblock Mutan: Multimodal tucker fusion for visual question answering.
\newblock In {\em ICCV}, 2017.

\bibitem{fashioniq-arxiv-2019}
Xiaoxiao Guo, Hui Wu, Yupeng Gao, Steven Rennie, and Rogerio Feris.
\newblock The fashion iq dataset: Retrieving images by combining side
  information and relative natural language feedback.
\newblock In {\em ICCV}, 2019.

\bibitem{he-arxiv-2015}
Kaiming He, Xiangyu Zhang, Shaoqing Ren, and Jian Sun.
\newblock {Deep Residual Learning for Image Recognition}.
\newblock In {\em CVPR}, 2016.

\bibitem{huang-cvpr-2017}
Gao Huang, Zhuang Liu, Laurens Van Der~Maaten, and Kilian~Q Weinberger.
\newblock Densely connected convolutional networks.
\newblock In {\em CVPR}, 2017.

\bibitem{Sergey-icml-2015}
Sergey Ioffe and Christian Szegedy.
\newblock {Batch Normalization: Accelerating Deep Network Training by Reducing
  Internal Covariate Shift}.
\newblock In {\em ICML}, 2015.

\bibitem{kingma-iclr-2015}
Diederik Kingma and Jimmy Ba.
\newblock {Adam: A Method for Stochastic Optimization}.
\newblock In {\em ICLR}, 2015.

\bibitem{li2019designovels}
Jianri Li, Jae whan Lee, Woo sang Song, Ki young Shin, and Byung hyun Go.
\newblock Designovel's system description for fashion-iq challenge 2019.
\newblock {\em arXiv}, 2019.

\bibitem{liu-cvpr-2016}
Ziwei Liu, Ping Luo, Shi Qiu, Xiaogang Wang, and Xiaoou Tang.
\newblock Deepfashion: Powering robust clothes recognition and retrieval with
  rich annotations.
\newblock In {\em Proceedings of IEEE Conference on Computer Vision and Pattern
  Recognition (CVPR)}, 2016.

\bibitem{miech-arxiv-2017}
Antoine Miech, Ivan Laptev, and Josef Sivic.
\newblock Learnable pooling with context gating for video classification.
\newblock {\em arXiv}, 2017.

\bibitem{Pennington-emnlp-2014}
Jeffrey Pennington, Richard Socher, and Christopher~D. Manning.
\newblock {GloVe: Global Vectors for Word Representation}.
\newblock In {\em EMNLP}, 2014.

\bibitem{srivastava-jmlr-2014}
Nitish Srivastava, Geoffrey Hinton, Alex Krizhevsky, Ilya Sutskever, and Ruslan
  Salakhutdinov.
\newblock {Dropout: A Simple Way to Prevent Neural Networks from Overfitting}.
\newblock {\em JMLR}, 2014.

\bibitem{sun-neurips-2018}
Shuyang Sun, Jiangmiao Pang, Jianping Shi, Shuai Yi, and Wanli Ouyang.
\newblock {FishNet: A Versatile Backbone for Image, Region, and Pixel Level
  Prediction}.
\newblock In {\em NeurIPS}, 2018.

\bibitem{vo-cvpr-2019}
Nam Vo, Lu Jiang, Chen Sun, Kevin Murphy, Li-Jia Li, Li Fei-Fei, and James
  Hays.
\newblock Composing text and image for image retrieval-an empirical odyssey.
\newblock In {\em CVPR}, 2019.

\bibitem{yu2020curlingnet}
Youngjae Yu, Seunghwan Lee, Yuncheol Choi, and Gunhee Kim.
\newblock Curlingnet: Compositional learning between images and text for
  fashion iq data.
\newblock {\em arXiv}, 2020.

\end{thebibliography}
}

\end{document}